\documentclass[letterpaper, 10 pt, conference]{ieeeconf}  

\IEEEoverridecommandlockouts                              

\title{\LARGE \bf Multi-Rate Nonlinear Model Predictive Control for Wall-Supported Bipedal Locomotion of Quadrupedal Robots}

\usepackage{amsmath,amssymb,amsfonts}
\usepackage{cite}
\usepackage{graphicx} 
\usepackage{epsfig} 
\usepackage{times} 
\usepackage{xcolor}
\usepackage{balance}
\usepackage{multicol}
\usepackage{array}
\usepackage{multirow}
\usepackage{booktabs}
\usepackage{xcolor}
\usepackage[colorlinks=true, allcolors=blue]{hyperref}

\newcommand{\Real}{\mathbb{R}}
\newcommand{\col}{\textrm{col}}
\newcommand{\Integer}{\mathbb{Z}_{\geq0}}
\newcommand{\diag}{\textrm{diag}}
\newcommand{\skews}{\mathbb S}
\newcommand{\identity}{\mathbb I}
\newcommand{\des}{\textrm{des}}

\newcommand{\SRB}{\textrm{SRB}}

\newcommand{\net}{\textrm{net}}
\newcommand{\SO}{\textrm{SO}(3)}
\newcommand{\so}{\mathfrak{so}(3)}
\newcommand{\terminal}{\textrm{terminal}}
\newcommand{\stage}{\textrm{stage}}
\newcommand{\reff}{\textrm{ref}}

\newcommand{\Xset}{\mathcal{X}}
\newcommand{\Uset}{\mathcal{U}}
\newcommand{\Vset}{\mathcal{V}}
\newcommand{\Tset}{\mathcal{T}}
\newcommand{\foot}{\textrm{foot}}
\newcommand{\ex}{\textrm{e}}

\usepackage{etoolbox}

\AtBeginEnvironment{equation}{\vspace{-3pt}}
\AtEndEnvironment{equation}{\vspace{-3pt}}
\AtBeginEnvironment{align}{\vspace{-3pt}}
\AtEndEnvironment{align}{\vspace{-3pt}}
\AtBeginEnvironment{alignat}{\vspace{-3pt}}
\AtEndEnvironment{alignat}{\vspace{-3pt}}

\newtheorem{example}{\textbf{Example}}

\newtheorem{problem}{\textbf{Problem}}

\author{Taizoon Chunawala$^{1}$, Jeeseop Kim$^{2}$, and Kaveh Akbari Hamed$^{1}$
\thanks{The work of T. Chunawala is partially supported by the National Science Foundation (NSF) under Grant 2306984. The work of K. Akbari Hamed is supported by the NSF under Grants 2024772 and 2423725.}
\thanks{$^{1}$T. Chunawala and K. Akbari Hamed are with the Department of Mechanical Engineering, Virginia Tech, Blacksburg, VA 24061, USA, {\tt\small \{taizoonac, kavehakbarihamed\}@vt.edu}}
\thanks{$^{2}$J. Kim is with the Department of Aerospace and Mechanical Engineering,
The University of Texas at El Paso, TX 79968, USA, {\tt\small jkim16@utep.edu}}
}

\begin{document}

\maketitle
\thispagestyle{empty}
\pagestyle{empty}


\begin{abstract}
This paper presents a novel layered planning and control framework based on multi-rate nonlinear model predictive control (MR-NMPC) that enables quadrupedal robots to perform hybrid bipedal locomotion with wall-assisted support in constrained environments. Real-time trajectory optimization for this locomotion presents significant challenges, as the controller must simultaneously plan for both the contact points and the continuous trajectories of the robot's center of mass (CoM) and orientation within the robot's nonlinear dynamics while accounting for unilateral contact constraints, underactuation, and the switching nature of the robot's dynamics. At the high level of the control framework, an MR-NMPC is proposed, which dynamically plans both the discrete-time trajectories of the contact points and the continuous-time trajectories of the CoM and orientation, using a single rigid body (SRB) dynamics model. By incorporating contact-point planning within the multi-rate optimal control framework, this approach enhances dynamic stability compared to heuristic foot placement strategies. At the low level of the control framework, a nonlinear whole-body controller (WBC) based on virtual constraints and a quadratic program enforces full-order dynamics and tracks the MR-NMPC references. The proposed approach is validated through extensive numerical simulations demonstrating the robust wall-assisted bipedal locomotion of a Unitree A1 quadrupedal robot on rough terrains and under external disturbances in a constrained environment. Comparative analysis shows that the proposed MR-NMPC achieves a 2.9 times higher success rate compared to conventional MPC with heuristic-based foot placement strategies in negotiating irregular terrain at high speeds.
\end{abstract}


\section{Introduction}
\label{sec:Introduction}

Bipedal locomotion in quadrupedal robots provides a means to extend their capabilities in environments where space is limited and mobility demands exceed the scope of traditional quadrupedal gaits. Operating in an upright posture allows quadrupeds to reach objects positioned at a higher height, negotiate obstacles that would obstruct four-legged motion, and perform manipulation tasks that require elevation of the torso. However, achieving fully dynamic bipedal balance in such constrained environments is especially challenging due to limited foothold options and the morphological design of quadrupeds. To address these challenges, we investigate \textit{wall-supported bipedal locomotion}, in which the robot leverages environmental contacts with vertical surfaces for additional stability. This strategy enables robots to operate effectively in tight corridors, cluttered industrial settings, and disaster sites, where leaning against walls or bracing for balance is often essential. For example, a quadrupedal robot navigating a narrow hallway can transition into a wall-supported bipedal posture to reach and operate a control panel that would otherwise be inaccessible in a purely quadrupedal stance. By exploiting walls as supportive contacts, quadrupeds gain a robust and practical mechanism for upright locomotion in constrained environments. At the same time, this strategy introduces a new set of challenges in modeling, control, and real-time execution, which we address in this work.

\begin{figure}
    \centering
    \includegraphics[width= 0.98\linewidth]{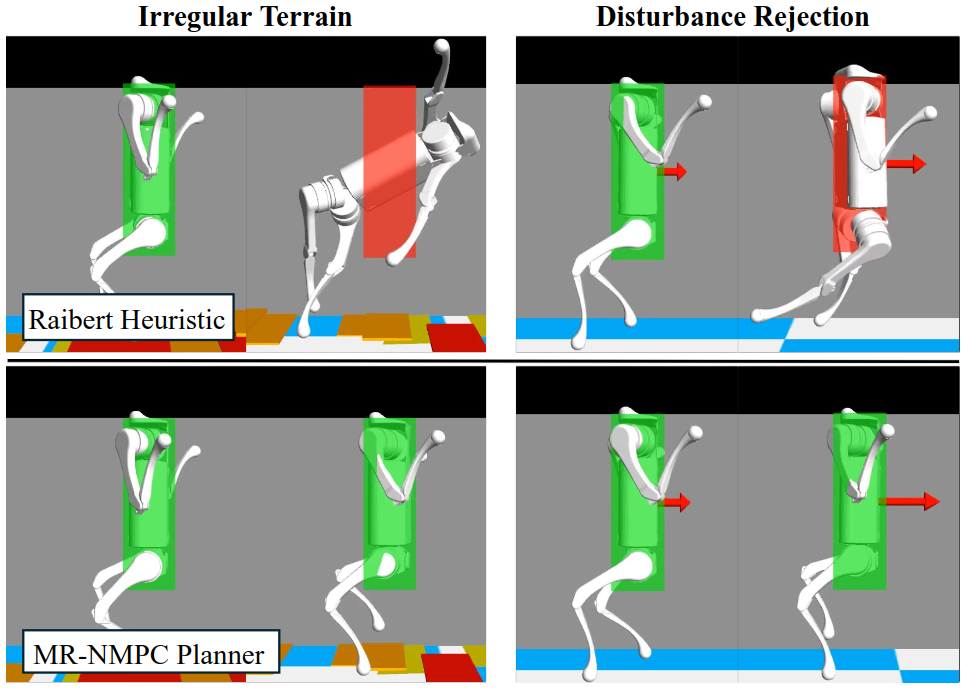}
    \vspace{-0.5em}
    \caption{Snapshots demonstrating wall-supported locomotion using the proposed MR-NMPC controller and the Raibert heuristic as a baseline. The green box denotes the prescribed safety envelope, while the red box indicates a violation of the safety envelope during the wall-supported locomotion.}
    \label{fig:illustration1}
    \vspace{-1.5em}
\end{figure}

Despite its potential, wall-supported bipedal locomotion introduces several challenges. First, the dynamics of quadrupedal robots in an upright posture are inherently hybrid, unstable, and high-dimensional, involving fast-varying reaction forces between the limbs and the environment together with slower variations in foot placement and contact transitions for both upper and lower limbs. This \textit{multi-rate structure} complicates the design of real-time predictive controllers for force and footstep planning, since standard nonlinear model predictive control (NMPC) formulations typically assume uniform input rates. Second, operating in constrained environments requires tight coupling between locomotion and environmental support forces, demanding simultaneous regulation of the center of mass (CoM) trajectory, torso orientation, and wall interaction forces while respecting feasibility constraints such as torque limits, kinematic reachability, and friction cones. Finally, these requirements must be satisfied under stringent real-time constraints, necessitating optimization-based control frameworks that can reliably perform predictive planning and whole-body coordination at high frequency despite significant computational complexity. 

The \textit{overarching goal} of this paper is to develop a unified and computationally efficient multi-rate nonlinear model predictive control (MR-NMPC) framework that simultaneously plans interaction forces and footstep placements, enabling quadrupedal robots to achieve robustly stable wall-supported bipedal locomotion in constrained environments.

\subsection{Related Work}
\label{sec:Related_work}

Model predictive control (MPC) has emerged as a powerful framework for trajectory optimization and feedback control in legged robots \cite{Patrick_TRO_Review}. A common strategy is to pair MPC with reduced-order, or template, models \cite{Full_Koditschek_Template}, which capture the essential features of locomotion while abstracting away the full complexity of high-dimensional robot dynamics. Among the most influential examples is the linear inverted pendulum (LIP) model \cite{kajita19991LIP}, together with its numerous extensions, including the angular momentum LIP \cite{ALIP}, the spring-loaded inverted pendulum (SLIP) \cite{SLIP}, variable-height inverted pendulum model \cite{Variable_height_inverted_pendulum}, the vertical LIP \cite{vLIP_Sreenath}, and the hybrid LIP \cite{HLIP_Ames}. Beyond pendulum-based abstractions, alternative reduced-order formulations such as centroidal dynamics \cite{orin2013centroidal} and the single rigid body (SRB) model \cite{Kim_Wensing_Convex_MPC_01,Wensing_VBL_HJB,Abhishek_Hae-Won_TRO,pandala2022robust,Leila_Hamed_RAL} have also been widely adopted, offering computationally tractable yet more physically representative descriptions of legged locomotion. While quadratic programming (QP)-based MPC formulations for linearized template models are highly efficient, they often struggle to capture the nonlinear effects inherent in models such as centroidal dynamics and SRB, limiting their ability to fully exploit the dynamics of legged systems. To overcome these limitations, NMPC has been employed for real-time trajectory planning of dynamic gaits and for safe motion in cluttered environments, see, e.g., \cite{NMPC_Park_02,Sleiman_RAL,NMPC_CBF_Ames_Hutter,NMPC_CBF_Sreenath,Hutter_anymal_cbf_inWBC,Basit_ASME,Basit_RAL,Patrick_TRO_Review}. In addition, several efficient methods have been proposed for whole-body NMPC, such as \cite{whole_body_NMPC_RAL,crocoddyl,BiConMP}, typically formulated as a nonlinear program (NLP). However, these approaches have been developed primarily for quadrupedal gaits in quadrupedal robots and have not been extended to wall-supported bipedal locomotion, where multi-rate interactions with the environment introduce fundamentally different control challenges.

Bipedal gaits of quadrupedal robots have been investigated primarily through reinforcement learning (RL)-based control approaches, which have demonstrated impressive performance in both simulation and hardware. For example, \cite{Multi_modal_locomotion_Chen} proposed a multi-modal locomotion framework that combines a hand-engineered transition strategy with a learning-based controller. In \cite{Learning-agile_locomotion_Laura}, a transfer learning framework was introduced to enable quadrupedal robots to jump or walk on their hind legs. An RL-based bi-level framework was developed in \cite{Learning_agile_bipedal_on_quadruped_Yunfei} to achieve agile, human-like bipedal locomotion. \cite{Leveraging_symmetry_Zhi} examined the benefits of leveraging symmetry in model-free RL to improve gait robustness, while \cite{Learning_bipedal_on_quadruped} introduced an adversarial motion priors-based method for bipedal locomotion, validated in simulation. Most recently, \cite{Diverse_locomotion_Barrier_Haewon} presented a learning framework with barrier-based style rewards that enabled quadrupedal, tripod, and bipedal gaits across legged robots. In addition, an RL-based algorithm was developed in \cite{yoneda2025kleiynquadrupedrobot} to endow the quadrupedal robot KLEIYN with wall-climbing capabilities.

Beyond learning-based methods, MPC-based approaches have also been explored for bipedal locomotion of quadrupedal robots. For instance, \cite{Semini_ADMM} proposed a distributed MPC algorithm based on the alternating direction method of multipliers (ADMM) to generate bipedal gaits. More recently, \cite{Contact_Implicit_MPC_HaeWon} introduced a contact-implicit MPC framework to enable front-leg rearing motions in quadrupedal robots.

Compared to free bipedal locomotion, wall-supported bipedal walking for quadrupedal robots introduces unique challenges, as it involves multi-contact gaits composed of four-contact, three-contact, and double-contact phases. In this setting, the rear legs enable bipedal locomotion, while the front legs exert forces against nearby walls to provide additional stability and propulsion. This renders the system inherently hybrid, with unilateral constraints acting on both the front and rear limbs. To address these challenges, we aim to design a unified MR-NMPC algorithm that simultaneously plans the optimal state trajectory, reaction forces, and foot placements for both front and rear legs.

\begin{figure*}[ht!]
    \centering
    \includegraphics[width=\textwidth]{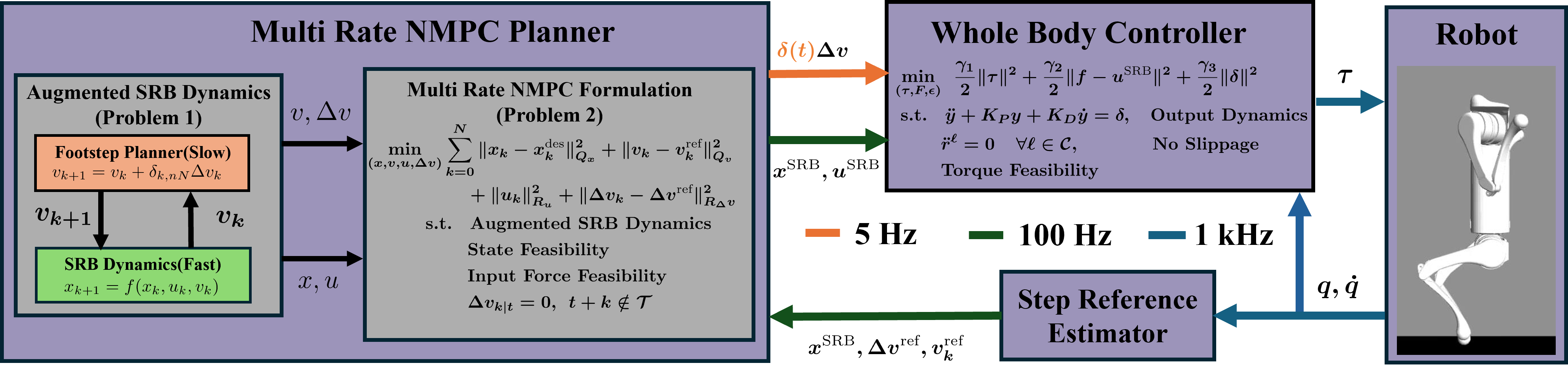}
    \vspace{-1.5em}
    \caption{Proposed layered control architecture: The high-level MR-NMPC optimizes reduced-order trajectories and footstep placements, while the low-level WBC (1 kHz) tracks these optimal values using full-order nonlinear dynamics and high-frequency sensor feedback.}
    \vspace{-1.5em}
    \label{fig:control_flow}
\end{figure*}

\subsection{Contributions}
\label{sec:Contributions}

This paper presents a layered optimal control framework for wall-supported bipedal locomotion in quadrupedal robots (Fig.~\ref{fig:illustration1}). The main contributions are:

\begin{itemize}
    \item A multi-rate nonlinear model predictive controller (MR-NMPC) that jointly optimizes continuous-time CoM motion, torso orientation, and environment reaction forces (ERFs), together with discrete-time contact-point trajectories for foot placement using a single rigid-body dynamics model.
    
    \item A multi-rate formulation that incorporates foothold evolution into the prediction dynamics by propagating contact-point locations across successive gait domains using step-length control inputs.
    
    \item A nonlinear whole-body controller (WBC), formulated as a convex quadratic program with virtual constraints, that tracks the optimized state, ERF, and contact-point trajectories at 1~kHz while enforcing the full-order robot dynamics.
    
    \item Validation in simulation on a Unitree A1 robot demonstrating wall-supported bipedal locomotion over rough terrain and under external disturbances, with comparisons against a heuristic Raibert-based foot-placement controller.
\end{itemize}


\section{High-Level MR-NMPC}
\label{sec:Gigh_Level_MR_NMPC}

This section establishes the mathematical foundation of the proposed high-level MR-NMPC framework (see Fig. \ref{fig:control_flow}). We formulate the framework for general nonlinear discrete-time dynamical systems that originate from template models of locomotion. In particular, we consider a multi-rate discrete-time system $\Sigma$ with two sets of control inputs, defined as
\begin{equation}\label{eq:state_eq}
    \Sigma: \quad x(t+1) = f\left(x(t),u(t),v(t)\right),
\end{equation}
where $t \in \Integer := \{0,1,\dots\}$ denotes the discrete-time index, $x(t) \in \Xset \subset \Real^{n_{x}}$ is the state vector, $u(t) \in \Uset \subset \Real^{n_{u}}$ is the \textit{fast} control input, and $v(t) \in \Vset \subset \Real^{n_{v}}$ is the \textit{slow} control input. The sets $\Xset$, $\Uset$, and $\Vset$ represent the feasible states, admissible fast inputs, and admissible slow inputs, respectively. 

The multi-rate structure arises from the fact that the fast input $u(t)$ can be updated at every time step, whereas the slow input $v(t)$ is allowed to change only at a predetermined subset of \textit{switching instants}, denoted by
\begin{equation}\label{eq:switching_times}
    \Tset := \{0 < t_{0} < t_{1} < t_{2} < \cdots\} \subset \Integer.
\end{equation}
Between these switching times, the slow input remains constant, yielding a piecewise-constant profile. Formally,  
\begin{equation}\label{eq:piecewise_constant}
    v(t+1) = v(t), \quad \text{if } t+1 \notin \Tset.
\end{equation}

\begin{example}[SRB Dynamics]\label{example:SRB_dynamics}
We adopt the SRB template model for wall-supported bipedal locomotion of quadrupedal robots. The state variables of the SRB dynamics are defined as the CoM position and orientation, together with their time derivatives,
\begin{equation}\label{eq:SRB_states}
    x:=\col(p,\dot{p},\theta,\omega)\in\Real^{n_{x}},\quad n_{x}=12,
\end{equation}
where $\col(\cdot)$ denotes the column operator, $p \in \Real^{3}$ represents the Cartesian coordinates of the center of mass (CoM), $\theta \in \Real^{3}$ denotes the Euler angles (roll, pitch, yaw), and $\omega \in \Real^{3}$ represents the angular velocity in the body frame. The corresponding rotation matrix is denoted by $R(\theta) \in \SO$.

In this work, we use the term environment reaction forces (ERFs) to denote the contact forces exerted by the environment on the robot. Unlike the conventional GRFs, ERFs include both ground contact forces and additional reaction forces arising from wall contacts during wall-supported locomotion. The fast control inputs $u(t)$ are the ERFs that generate the net force and torque about the CoM, denoted by $(f^{\net}, \tau^{\net}) \in \Real^{3}$ and expressed in the world frame. The SRB dynamics are governed by
\begin{equation}\label{eq:SRB_dyn}
    \Sigma^{\SRB}: \begin{cases}
    \ddot{p}      = \frac{f^{\net}}{m} - g_0 \\
    \dot{\theta}  = A(\theta)\,\omega \\
    \dot{\omega}  = I^{-1} \left(R^\top(\theta)\, \tau^{\net} - \skews(\omega) \, I\, \omega \right),
\end{cases}
\end{equation}
where $m$ denotes the total mass of the quadrupedal robot, $g_{0} \in \Real^{3}$ is the gravitational vector, $I \in \Real^{3\times3}$ is the inertia matrix expressed in the body frame, and $\skews(\cdot): \Real^3 \rightarrow \so$ is the skew-symmetric matrix operator, satisfying $\skews(a)\,b = a \times b$ for all $a,b \in \Real^{3}$. The transformation matrix $A(\theta) \in \Real^{3\times3}$ maps the body angular velocity to the time derivative of the Euler angles.

The net force and torque acting on the CoM are computed as
\begin{equation}
    \begin{bmatrix}
    f^{\net}\\
    \tau^{\net}
    \end{bmatrix}:=\sum_{\ell\in\mathcal{C}} \begin{bmatrix}
        f^{\ell}\\
        \skews(r^{\ell})\,f^{\ell}
    \end{bmatrix},
\end{equation}
where $\ell\in\mathcal{C}$ indexes the stance feet, $\mathcal{C}\subseteq\{1,\cdots,4\}$ denotes the set of stance feet, $f^{\ell}\in\Real^{3}$ is the ERF at stance foot $\ell$, and $r^{\ell}\in\Real^{3}$ is the vector from CoM to foot $\ell$,
\begin{equation}
r^{\ell} := r^{\foot,\ell} - p,
\end{equation}
with $r^{\foot,\ell}$ denoting the Cartesian coordinates of foot $\ell$ in the world frame. 


To facilitate a unified NMPC framework that co-optimizes both ERFs and foot placements for wall-supported locomotion, we augment the state vector to include the Cartesian coordinates of the stance limb end-effectors, $r^{\foot,\ell}$. In this formulation, the foot positions are treated as states that evolve according to the ``slow'' control inputs $\Delta v(t)$, which represent the incremental step lengths:
\begin{equation}\label{eq:foot_evolution}
\Delta {r}^{\foot,\ell} = \Delta v^{\ell}(t), \quad \ell \notin \mathcal{C}
\end{equation}
By integrating foot positions into the state space, the NMPC can directly reason about the coupling between SRB dynamics and future support configurations, enabling agile maneuvers over unknown terrain.

The equations of motion in \eqref{eq:SRB_dyn} are discretized using Euler’s method and expressed in the multi-rate nonlinear state-space form \eqref{eq:state_eq}. In this formulation, the ERFs can vary at every time step and are therefore treated as fast control inputs, whereas the limb placements are updated only at the beginning of each gait cycle and are regarded as slow inputs. Moreover, the switching times naturally correspond to the beginning of each gait cycle, with the piecewise-constant property in \eqref{eq:piecewise_constant} enforcing continuity of the stance limb coordinates within a cycle. This example demonstrates how the SRB template model integrates seamlessly into the general multi-rate NMPC framework introduced above, where both fast and slow inputs are optimized in a \textit{unified manner}.
\end{example}

Returning to the general formulation in \eqref{eq:state_eq}, we are interested in the following problem.

\begin{problem}[Augmented SRB Dynamics]\label{problem:steering}
We aim to design a unified, computationally efficient, and real-time MR-NMPC algorithm for a predetermined set of switching instants $\Tset$ that computes the pair of optimal fast and slow control inputs $(u(t),v(t))$ to steer the nonlinear dynamics \eqref{eq:state_eq} from an initial condition $x(0) \in \Xset$ toward a reference trajectory $x^{\mathrm{ref}}(t) \in \Xset$, while ensuring $x(t) \in \Xset$, $u(t) \in \Uset$, $v(t) \in \Vset$, and enforcing the piecewise-constant property \eqref{eq:piecewise_constant} for all $t \in \Integer$.
\end{problem}

Problem~\ref{problem:steering} is formulated for the original nonlinear, time-invariant system \eqref{eq:state_eq} with both fast and slow inputs. However, enforcing the piecewise-constant behavior of the slow input $v(t)$ in \eqref{eq:piecewise_constant} introduces significant challenges for real-time trajectory planning. To overcome this difficulty, we reformulate the problem using an indicator-function representation. This approach transforms the original steering problem into an equivalent formulation for a time-varying augmented nonlinear system with augmented inputs that are allowed to update at every time step, in contrast to the original formulation. 

Specifically, we define the \textit{indicator function} $\delta_{\Tset} : \Integer \rightarrow \{0,1\}$ as
\begin{equation}\label{eq:indicator_func}
    \delta_{\Tset}(t) :=
    \begin{cases}
        1, & t \in \Tset \\
        0, & t \notin \Tset,
    \end{cases}
\end{equation}
which specifies whether the current time $t$ corresponds to a switching instant. The evolution of the slow input can then be written as
\begin{equation}\label{eq:v_dynamics}
    v(t+1) = v(t) + \delta_{\Tset}(t+1)\,\Delta v(t),
\end{equation}
where $\Delta v(t) \in \Real^{n_{v}}$ denotes the update applied at switching instants. Thus, $v(t)$ remains constant between updates and changes only when $\delta_{\Tset}(t+1) = 1$. From \eqref{eq:v_dynamics}, the trajectory of the slow input can be explicitly written as
\begin{equation}\label{eq:v_closed_form}
v(t) = v(0) + \sum_{k=0}^{t-1} \delta_{\Tset}(k+1)\,\Delta v(k), \quad \forall t \geq 1,
\end{equation}
where $v(0)$ denotes the initial value of the slow input.

Incorporating \eqref{eq:state_eq} and \eqref{eq:v_dynamics} yields the augmented state model
\begin{equation}\label{eq:aug_dynamics}
    \Sigma^{a}:\;\begin{cases}
        x(t+1) = f\!\left(x(t),u(t),v(t)\right), \\[6pt]
        v(t+1) = v(t) + \delta_{\Tset}(t+1)\,\Delta v(t),
    \end{cases}
\end{equation}
where the pair $x^{a}(t):=\col(x(t),v(t)) \in \Xset \times \Vset$ defines the augmented state and $u^{a}(t):=\col(u(t),\Delta v(t)) \in \Uset \times \Real^{n_{v}}$ defines the augmented control input. Since the indicator function $\delta_{\Tset}(t)$ vanishes outside the switching set $\Tset$, both $(u(t),\Delta v(t))$ can be interpreted as fast inputs in this reformulated system. In particular, the evolution of $v(t)$ is effectively gated by $\delta_{\Tset}(t)$, ensuring that updates occur only at switching instants. For the SRB dynamics in Example \ref{example:SRB_dynamics}, the augmented state is formed by stacking the CoM positions, CoM velocities, Euler angles, angular velocities, and Cartesian coordinates of the stance limbs’ end-effectors. We now present the following problem, which is equivalent to the original Problem~\ref{problem:steering}.

\begin{problem}[MR-NMPC Formulation]\label{problem:augmented_steering}
We aim to design a computationally efficient and real-time NMPC algorithm for a predetermined set of switching instants $\Tset$ that computes the pair of optimal inputs $(u(t),\Delta v(t))$, both of which can be updated at every time step, to steer the augmented nonlinear time-varying dynamics \eqref{eq:aug_dynamics} from an initial condition $x^{a}(0) := \col(x(0),v(0)) \in \Xset \times \Vset$ toward the augmented reference trajectory $x^{a,\mathrm{ref}}(t) := \col(x^{\mathrm{ref}}(t),v^{\mathrm{ref}}(t)) \in \Xset \times \Vset$, while ensuring that $x(t) \in \Xset$, $u(t) \in \Uset$, and $v(t) \in \Vset$ for all $t \in \Integer$.
\end{problem}

To address Problem \ref{problem:augmented_steering}, we propose the following real-time and unified NMPC
\begin{alignat}{4}
&\min_{(x(\cdot),v(\cdot),u(\cdot),\Delta v(\cdot))} &&\mathcal{L}_{\terminal,x}\left(x_{t+N|t}\right) + \mathcal{L}_{\terminal,v}\left(v_{t+N|t}\right) \nonumber\\
& &&+ \sum_{k=0}^{N-1} \mathcal{L}_{\stage,x}\left(x_{t+k|t},u_{t+k|t}\right) \nonumber\\
& &&+ \sum_{k=0}^{N-1} \mathcal{L}_{\stage,v}\left(v_{t+k|t},\Delta v_{t+k|t}\right) \nonumber\\
& \hspace{1cm} \textrm{s.t.} && \!\!\!\!\!x_{t+k+1|t} = f\left(x_{t+k|t},u_{t+k|t},v_{t+k|t}\right) \nonumber\\
& &&\!\!\!\!\! v_{t+k+1|t} = v_{t+k|t} + \delta_{\Tset}(t+k+1)\,\Delta v_{t+k|t} \nonumber\\
& &&\!\!\!\!\! E_{u}(t+k)\,u_{t+k|t} =0 \nonumber\\
& &&\!\!\!\!\! E_{\Delta v}(t+k)\,\Delta v_{t+k|t} =0 \nonumber\\
& &&\!\!\!\!\! x_{t+k|t} \in \Xset, \quad u_{t+k|t} \in \Uset, \quad v_{t+k|t} \in \Vset,\label{eq:MR_NMPC}
\end{alignat}
where $N$ denotes the control horizon. The variables $x_{t+k|t}$, $u_{t+k|t}$, $v_{t+k|t}$, and $\Delta v_{t+k|t}$ represent the predicted state, fast input, slow input, and slow-input update at stage $k$, computed at time $t$ using the augmented prediction model \eqref{eq:aug_dynamics}. The initial condition is set by the measured variables at time $t$, namely $x_{t|t}=x(t)$ and $v_{t|t}=v(t)$. In addition, $x(\cdot)$, $u(\cdot)$, $v(\cdot)$, and $\Delta v(\cdot)$ denote the corresponding trajectories over the control horizon. The cost function consists of the terminal and stage costs for both $x$ and $v$ variables, defined as $\mathcal{L}_{\terminal,x} := \|x_{t+N|t} - x_{t+N|t}^{\reff}\|_{P_{x}}^{2}$, $\mathcal{L}_{\terminal,v} := \|v_{t+N|t} - v_{t+N|t}^{\reff}\|_{P_{v}}^{2}$, and $\mathcal{L}_{\stage,x} := \|x_{t+k|t} - x^{\reff}_{t+k|t}\|_{Q_{x}}^{2} + \|u_{t+k|t}\|_{R_{u}}^{2}$, $\mathcal{L}_{\stage,v}:= \|v_{t+k|t} - v^{\reff}_{t+k|t}\|_{Q_{v}}^{2} + \|\Delta v_{t+k|t}-\Delta v^{\textrm{ref}}\|_{R_{\Delta v}}^{2}$
where $P_{x}$, $P_{v}$, $Q_{x}$, $Q_{v}$, $R_{u}$, and $R_{\Delta v}$ are positive-definite weighting matrices. Notably, the framework utilizes steplength obtained by Raibert heuristic as $\Delta v^{\textrm{ref}}$ and the corresponding foot evolution as $v^{\textrm{ref}}_{*}$. Throughout this paper, we use the notation $\|z\|_{Q}^{2} := z^\top Q z$ for any vector $z$.  

The equality constraints of the NMPC are induced by the nonlinear, time-varying augmented model \eqref{eq:aug_dynamics}, together with the assignment conditions $E_{u}(t+k)\,u_{t+k|t}=0$ and $E_{\Delta v}(t+k)\,\Delta v_{t+k|t}=0$, which enforce availability of the control inputs depending on the contact configuration. For example, in the SRB dynamics, the sequence of contact points with the environment changes over the control horizon, which restricts the admissible ERFs or foot placements at each stage $t+k$. Here, $E_{u}(t+k)$ and $E_{\Delta v}(t+k)$ are time-varying binary matrices (composed of zeros and ones) that constrain unavailable inputs to zero. The inequality constraints originate from state feasibility and input admissibility. 

\begin{example}[Locomotion Pattern]
For the wall-supported bipedal locomotion of the quadrupedal robot, we consider a multi-contact locomotion pattern denoted by $\mathcal{G}$. The periodic gait cycle is modeled as a sequence of three phases: a four-contact phase, followed by a three-contact phase, and concluding with a double-contact phase (see Fig. \ref{fig:contactschedule}). The durations of the four-contact, three-contact, and double-contact phases are set to 50 ms, 30 ms, and 120 ms, respectively. The sampling time used to discretize the SRB dynamics is $T_{s}=10$ ms, corresponding to solving the MR-NMPC at 100 Hz. The control horizon is chosen as $N=20$ samples, which corresponds to a prediction window of $N T_{s}=200$ ms. Consequently, the MR-NMPC optimizes trajectories across multiple phases of the gait cycle. For example, when the NMPC is solved in the middle of the four-contact phase (for example at $30$ ms time step), the prediction horizon encompasses the current four-contact phase, the subsequent three-contact and double-contact phases of the same cycle, as well as part of four-contact phase (3 time steps) in the next cycle.

\begin{figure}[t!]
    \centering
    \includegraphics[width=0.9\linewidth]{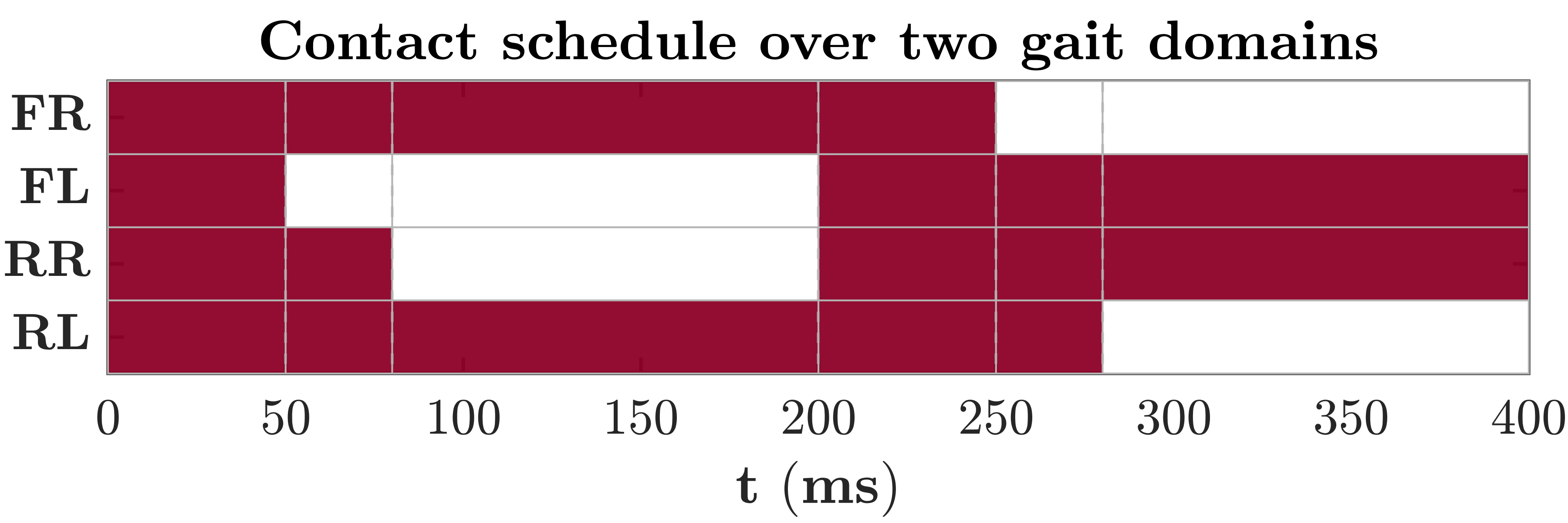}
    \vspace{-0.5em}
    \caption{Gait contact schedule over a 400 ms interval across two gait domains. Dark maroon regions denote the stance phase, while white regions indicate the swing phase.}
    \vspace{-1.5em}
    \label{fig:contactschedule}
\end{figure}

The resulting optimal CoM and orientation trajectories, together with the ERF trajectories (fast inputs) and limb end-effector trajectories (slow inputs), are provided to the low-level nonlinear whole-body controller (WBC) for tracking. The numerical implementation details and the solver used for the high-level MR-NMPC will be discussed in Section \ref{sec:Setup}.
\end{example}



\section{Low-Level Nonlinear WBC}
\label{sec:Low_Level_Nonlinear_WBC}

This section describes the low-level nonlinear whole-body controller (WBC) developed for the full-order dynamics of the quadrupedal robot during wall-supported bipedal locomotion. The controller is adapted from \cite{Randy_Paper_LCSS,pandala2022robust}, with modifications tailored to the proposed framework.

The full-order floating-base dynamics of the quadrupedal robot during wall-supported locomotion are expressed using the Euler–Lagrange equations and the principle of virtual work as
\begin{alignat}{4}
    & D(q)\,\ddot{q} + H(q,\dot{q}) = B\,\tau + \sum_{\ell\in\mathcal{C}} \left(J^{\ell}(q)\right)^\top f^{\ell} \label{eq:full_order_dyn}\\
    & \ddot{r}^{\foot,\ell} = 0, \quad \forall \ell \in \mathcal{C}, \label{eq:contact_conditions}
\end{alignat}
where $q \in \mathcal{Q} \subset \Real^{n_{q}}$ denotes the generalized coordinates, $\tau \in \mathcal{T} \subset \Real^{n_{\tau}}$ represents the joint torques, and $f^{\ell} \in \Real^{3}$ is the ERF applied at the stance limb end-effector $\ell \in \mathcal{C}$. The matrix $D(q) \in \Real^{n_{q}\times n_{q}}$ is the positive-definite mass–inertia matrix, $H(q,\dot{q}) \in \Real^{n_{q}}$ collects the Coriolis, centrifugal, and gravitational terms, $B \in \Real^{n_{q}\times n_{\tau}}$ is the input distribution matrix, and $J^{\ell}(q) \in \Real^{3\times n_{q}}$ denotes the contact Jacobian matrix associated with stance limb $\ell \in \mathcal{C}$. The contact constraints are enforced by requiring the stance limb end effector accelerations to vanish, as specified in \eqref{eq:contact_conditions}.

To bridge the gap between reduced- and full-order locomotion models, we define a set of outputs, referred to as virtual constraints \cite{Jessy_Book}, which impose the full-order model to track the optimal trajectories prescribed by the high-level MR-NMPC. Specifically, we define
\begin{equation}\label{eq:virtual_constraints}
y(t,q) := y_{a}(q) - y_{\des}(t),
\end{equation}
where $y_{a}(q)$ denotes the actual output variables to be controlled (controlled variables), and $y_{\des}(t)$ represents their desired evolution along the gait. The controlled variables include the CoM position, torso orientation (Euler angles), and the Cartesian coordinates of the swing limb end effectors. 
Optimal values $(x_{t+1|t}^{\star},u_{t|t}^{\star},v_{t+1|t}^{\star},\Delta v_{t|t}^{\star})$ from the SRB dynamics are used to construct the desired output function $y_{\des}(t)$. For the swing limb end effectors, we employ a Bézier polynomial that interpolates between the current position of the limb and the upcoming placement specified in $v_{t+1|t}$.

We next formulate the following real-time convex quadratic program (QP) to solve for the joint-level torques:
\begin{alignat}{4}
    & \min_{(\tau,f,\delta)} \hspace{0.4cm} && \frac{\gamma_{1}}{2} \|\tau\|^{2} + \frac{\gamma_{2}}{2} \|f-f_{\des}\|^{2} && + \frac{\gamma_{3}}{2} \|\delta\|^{2} \nonumber\\
    & \hspace{0.2cm} \textrm{s.t.} && \ddot{y} + K_{D}\, \dot{y} + K_{P}\, y = \delta && \textrm{(Output dynamics)} \nonumber\\
    & && \ddot{r}^{\foot,\ell} = 0,\quad \forall \ell \in \mathcal{C} && \textrm{(No slippage)} \nonumber\\
    & && \tau \in \mathcal{T}, \quad\,\, f \in \mathcal{FC} && \textrm{(Feasibility)},
    \label{eq:QP_WBC}
\end{alignat}
where $\gamma_{1}$, $\gamma_{2}$, and $\gamma_{3}$ are positive weighting factors, $f := \col\{f^{\ell} \mid \ell \in \mathcal{C}\}$ denotes the stacked vector of all ERFs at stance limbs, $f_{\des}$ represents the desired ERFs prescribed by the high-level MR-NMPC; $\delta$ is the slack variable that ensures feasibility of the output dynamics, and $\mathcal{FC}$ denotes the linearized friction cone constraints. The first equality constraint enforces the desired output dynamics, $\ddot{y} + K_{D}\,\dot{y} + K_{P}\,y = \delta$, where $K_{P}$ and $K_{D}$ are positive-definite gain matrices for output regulation, and $\delta$ is used to relax the dynamics when necessary. The second equality constraint imposes zero acceleration of the stance limb end effectors. Finally, the inequality constraints ensure the feasibility of the joint-level torques and stance-limb ERFs. The cost function minimizes a weighted sum of the squared 2-norms of the joint torques, the ERF tracking error, and the slack variables, with relative importance determined by the weighting factors $\gamma_{1}, \gamma_{2}, \gamma_{3}$. We remark that $\ddot{y}$ and $\ddot{r}^{\foot,\ell}$ are affine functions of $(\tau,f)$ under the Lagrangian dynamics \eqref{eq:full_order_dyn}; therefore, the optimization problem \eqref{eq:QP_WBC} is a convex QP, which will be reliably solved in real time at 1 kHz.


\begin{figure}
    \centering
    \includegraphics[width=0.97\linewidth]{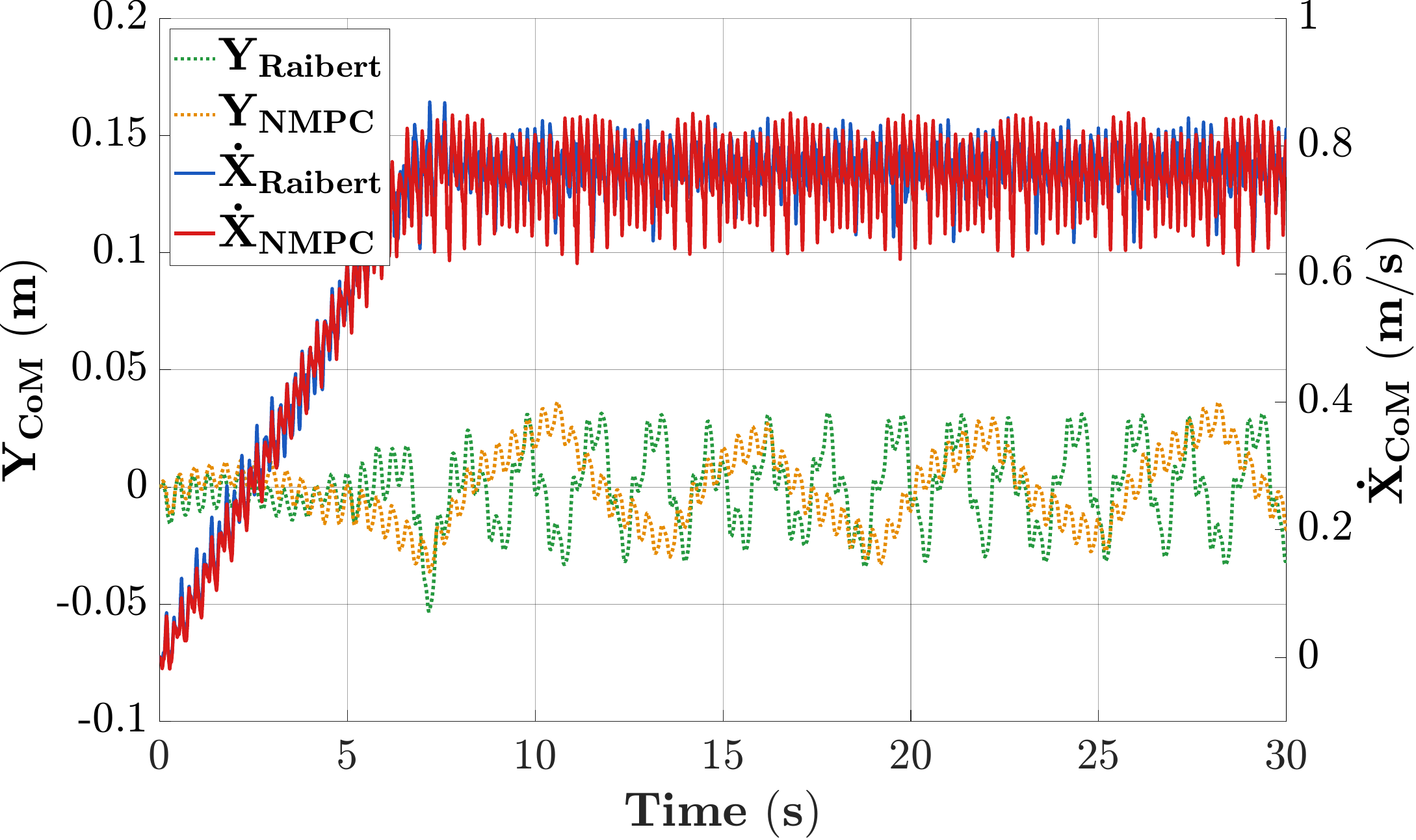}
    \vspace{-0.5em}
    \caption{A comparison of the CoM velocity and base orientation tracking for the Raibert heuristic (blue) and the proposed MR-NMPC (red) in a flat-terrain condition where the desired forward velocity is 0.8 m/s.}
    \vspace{-1.5em}
    \label{fig:raibvplan_flat}
\end{figure}

\section{Simulation Results}
\label{sec:Simulation}

To evaluate the proposed footstep planner, we conduct simulations of upright, wall-supported walking using a Unitree A1 quadruped within the RaiSim environment~\cite{RAISIM}. The simulation environment consists of a $45$ cm narrow corridor between parallel walls with unknown terrain featuring randomly placed wooden blocks similar to the one shown in Fig. \ref{fig:illustration1} to simulate a challenging, obstacle-laden path. 


\subsection{Setup and Controller Synthesis}
\label{sec:Setup}

This work employs the Unitree A1 quadruped for numerical validation of the proposed MR-NMPC framework. The $13$ kg A1 ($0.25$ m tall) features 18~DoFs, including 12 actuated leg joints (hip pitch, hip roll, and knee pitch). Numerical validation is performed using the RaiSim physics engine. The hyperparameters of the MR-NMPC for the SRB component are chosen as $Q^{\SRB}=\textrm{block diag}\{Q^{\SRB}_{p},Q^{\SRB}_{\dot{p}},Q^{\SRB}_{\theta},Q^{\SRB}_{\omega}\}$ with  $Q^{\SRB}_{p}=\diag\{1 \ex4,5\ex4,1\ex4\}$, $Q^{\SRB}_{\dot{p}}=\diag\{1\ex5,1\ex4,1\ex4\}$, $Q^{\SRB}_{\theta}=\diag\{8\ex4,8\ex5,3\ex4\}$, and $Q^{\SRB}_{\omega}=\diag\{1\ex2,1\ex2,1\ex2\}$. The terminal cost and control penalty are set to $P^{\SRB}=Q^{\SRB}$ and $R^{\SRB}= 0.01\,\identity_{12\times12}$, respectively.

For the footstep placement and step length optimization of the SRB model, the weighting matrices $Q_v$ and $R_{\Delta v}$ from are defined as block-diagonal structures $Q_v=\textrm{block diag}\{Q^{\textrm{FR}},Q^{\textrm{FL}},Q^{\textrm{RR}},Q^{\textrm{RL}}\}$, $R_{\Delta v}=\textrm{block diag}\{Q_v^{\textrm{FR}},Q_v^{\textrm{FL}},Q_v^{\textrm{RR}},Q_v^{\textrm{RL}}\}$. To ensure symmetric locomotion behavior, the weights are set uniformly across all four limbs with $Q^i=1\ex 3$ and $Q_v^i=1\ex 4$ for $i \in \{\textrm{FR},\textrm{FL},\textrm{RR},\textrm{RL}\}$. To maintain the focus on high-speed longitudinal agility, we reduce the dimensionality of the slow-input update $\Delta v$ by considering only the footstep position and step length along the x-direction of motion. Consequently, the decision variables for footstep planning are constrained to the sagittal plane.

The high-level MR-NMPC is implemented using the CasADi framework \cite{CasADI} with the IPOPT \cite{IPOPT} interior-point solver on a desktop PC (Intel Core i9-12900F, 64 GB RAM). The solver achieves a mean execution time of 8.26 ms with a standard deviation of 0.87 ms, which ensures that the high-level planning loop consistently operates at 100 Hz.


\begin{figure}[t!]
    \centering
    \includegraphics[width=0.97\linewidth]{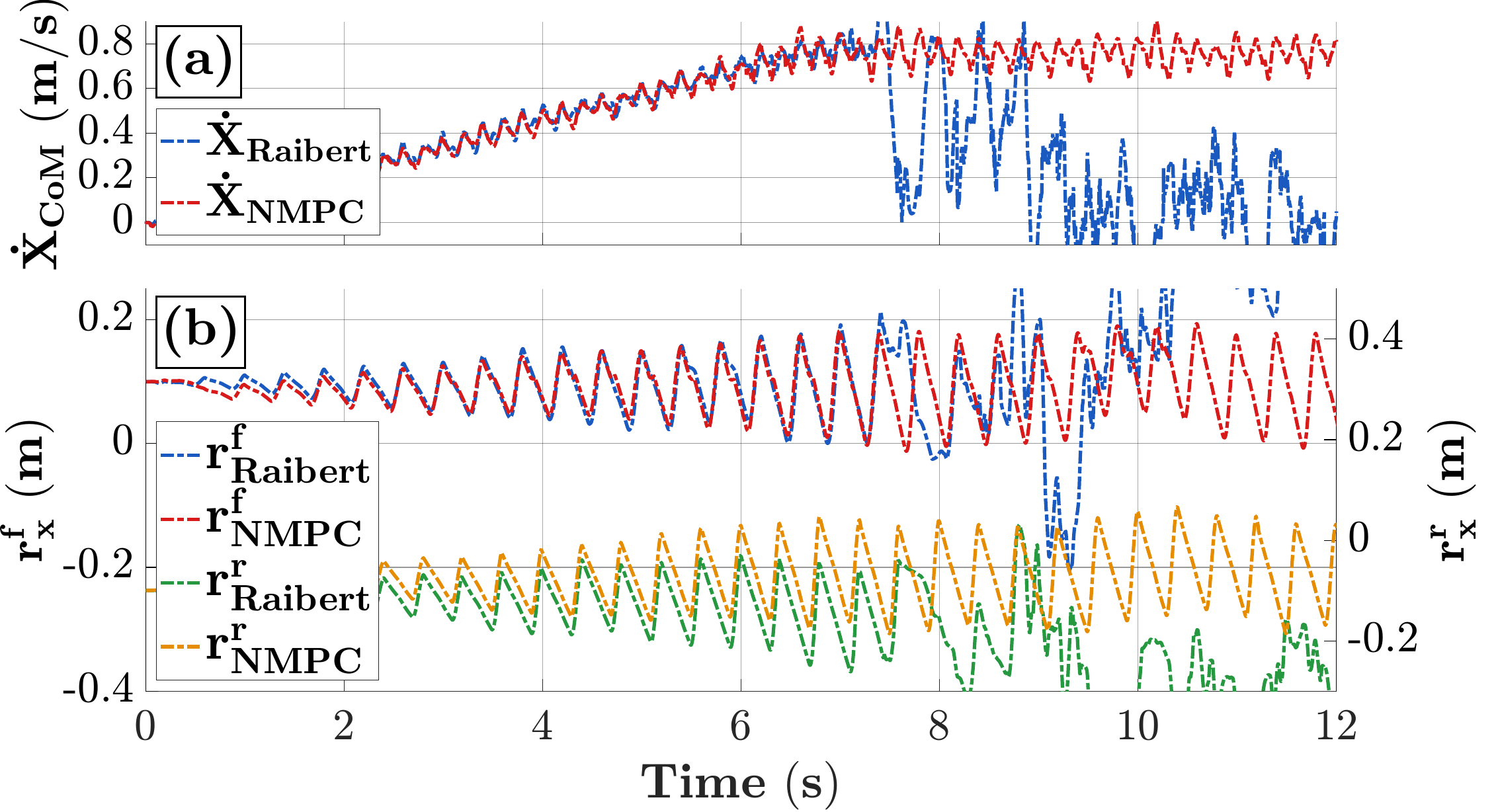}
    \vspace{-0.5em}
    \caption{Plots of the (a) CoM velocity with the Raibert heuristic (blue) and with MR-NMPC (red), where $0.8$ m/s is the desired velocity (b) front and rear foot position evolution with respect to the CoM position with the Raibert heuristic (blue) and with MR-NMPC (red).}
    \vspace{-1.25em}
    \label{fig:terrain_tracking}
\end{figure}

\subsection{Comparative Benchmark and Baseline Validation}

Previous MPC formulations optimize future footholds as independent variables or static parameters within the prediction horizon \cite{kang2022nonlinearmodelpredictivecontrol,9385863}. In contrast, the proposed MR-NMPC models foothold evolution as part of the prediction dynamics by propagating future foothold locations across successive gait domains using optimized step-length inputs.

To evaluate the benefits of this formulation, we benchmark the proposed controller against the classical Raibert heuristic \cite{eventbased_Raibert}, which computes the step adjustment as $\Delta x=\frac{\dot{x}T_s}{2}+k_v(\dot{x}-\dot{x}_{\mathrm{des}})$,
where $k_v=\sqrt{h/g}$.

Both controllers are first evaluated under nominal conditions (flat terrain with no external disturbances). As shown in Fig.~\ref{fig:raibvplan_flat}, the MR-NMPC and Raibert heuristic achieve nearly identical reference velocity ($v_{\mathrm{des}}=0.8$ m/s) and base orientation tracking, establishing a common baseline. This enables the performance gains in the subsequent disturbed and uneven-terrain scenarios to be attributed to the proposed predictive foothold formulation rather than controller tuning.



\subsection{Adaptive Footstep Placement and Stability}
\label{sec:adaptive_footstep}

The robustness of the proposed MR-NMPC framework during obstacle negotiation is evaluated by traversing a series of piled wooden blocks. The wooden blocks are $2$ cm tall, $14$ cm wide, stacked up to 3 layers and cover $70\%$ of the test track.  

The MR-NMPC utilizes full-state feedback and a 20-step prediction horizon to maintain the robot within the safety envelope and track the reference velocity as shown in Fig. \ref{fig:terrain_tracking} (a). To analyze the mechanism behind this stability, we examine the longitudinal foot offsets relative to the CoM, defined for the front ($r_x^f$) and rear ($r_x^r$) limbs as:
\begin{equation}
r^f_x = p^{\textrm{FR}}_x - x_{\textrm{CoM}}, \quad r^r_x = p^{\textrm{RR}}_x - x_{\textrm{CoM}}.
\end{equation}

As illustrated in Fig. \ref{fig:terrain_tracking} (b), the MR-NMPC diverges significantly from the Raibert heuristic in its treatment of the rear limbs. While the front foot placement ($r_x^f$) remains somewhat comparable between the two methods, the NMPC proactively modulates the rear foot placement ($r_x^r$) to accommodate the undulating surface. By decoupling the front and rear step lengths, the optimizer can dynamically expand or contract the support polygon. This flexibility allows the framework to generate corrective restorative torques that are unattainable under the fixed-symmetry constraints of the Raibert heuristic, effectively ``steering'' the robot back to the reference path despite the treacherous terrain.    

\begin{figure}[t!]
    \centering
    \includegraphics[width=0.92\linewidth]{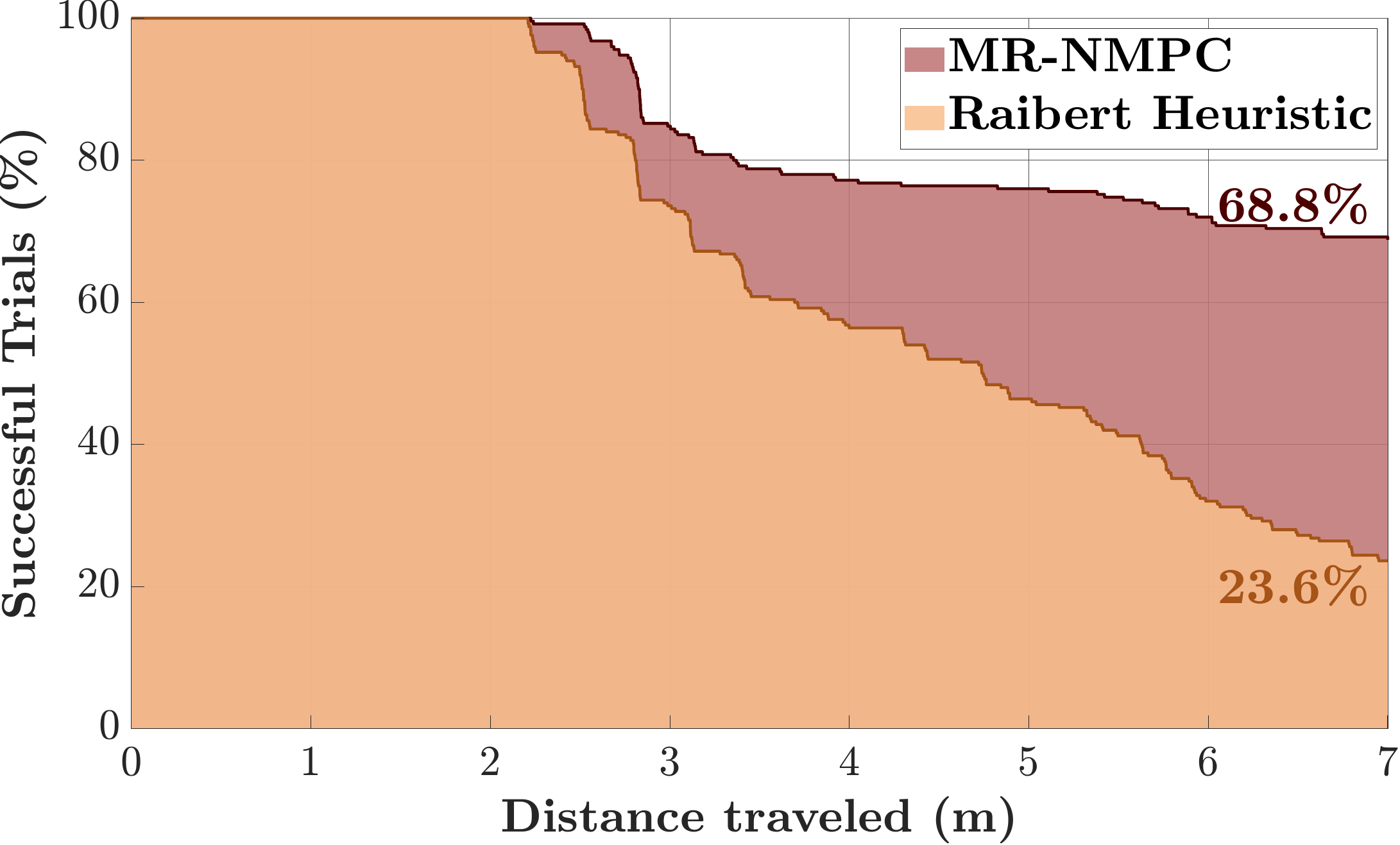}
    \vspace{-0.5em}
    \caption{Comparison of proposed MR-NMPC planner against Raibert heuristic over 250 randomly generated rough terrains.}
    \vspace{-1.0em}
    \label{fig:raibvplan}
\end{figure}

\subsection{Comparative and Quantitative Analysis}
\label{sec:Comparison}

We evaluate the proposed MR-NMPC against the Raibert baseline across 250 randomized $7$ m tracks featuring stochastic wooden block placement. One such terrain is shown in Fig. \ref{fig:illustration1}. The reference velocity is ramped up in steps of $0.05$ m/s every two gait domains to a maximum of $0.8$ m/s to test the algorithm across a wide speed regime. 
To maintain a rigorous benchmark, we define a successful trial based on a multi-modal stability envelope: nominal CoM height, lateral tracking margins, and kinematic safety limits. A trial is terminated upon the first constraint violation, and the traversal distance is recorded.

Figure \ref{fig:raibvplan} illustrates the survival probability across these 250 trials. While both planners demonstrate comparable reliability in the low-velocity regime (up to $2$ m of travel), the Raibert heuristic exhibits a sharp degradation in success rate as the reference velocity increases. This divergence highlights a fundamental limitation of velocity-based heuristics in agile locomotion: at higher speeds, the increased distance covered per stance phase necessitates a predictive look-ahead to manage the coupled base-and-footstep dynamics. By co-optimizing the full robot state—the MR-NMPC proactively adapts foothold locations based on the anticipated contact configuration to allow the system to navigate irregular terrain at high speeds where reactive feedback alone is insufficient.

 \begin{figure}[t!]
\centering
\includegraphics[width=0.97\linewidth]{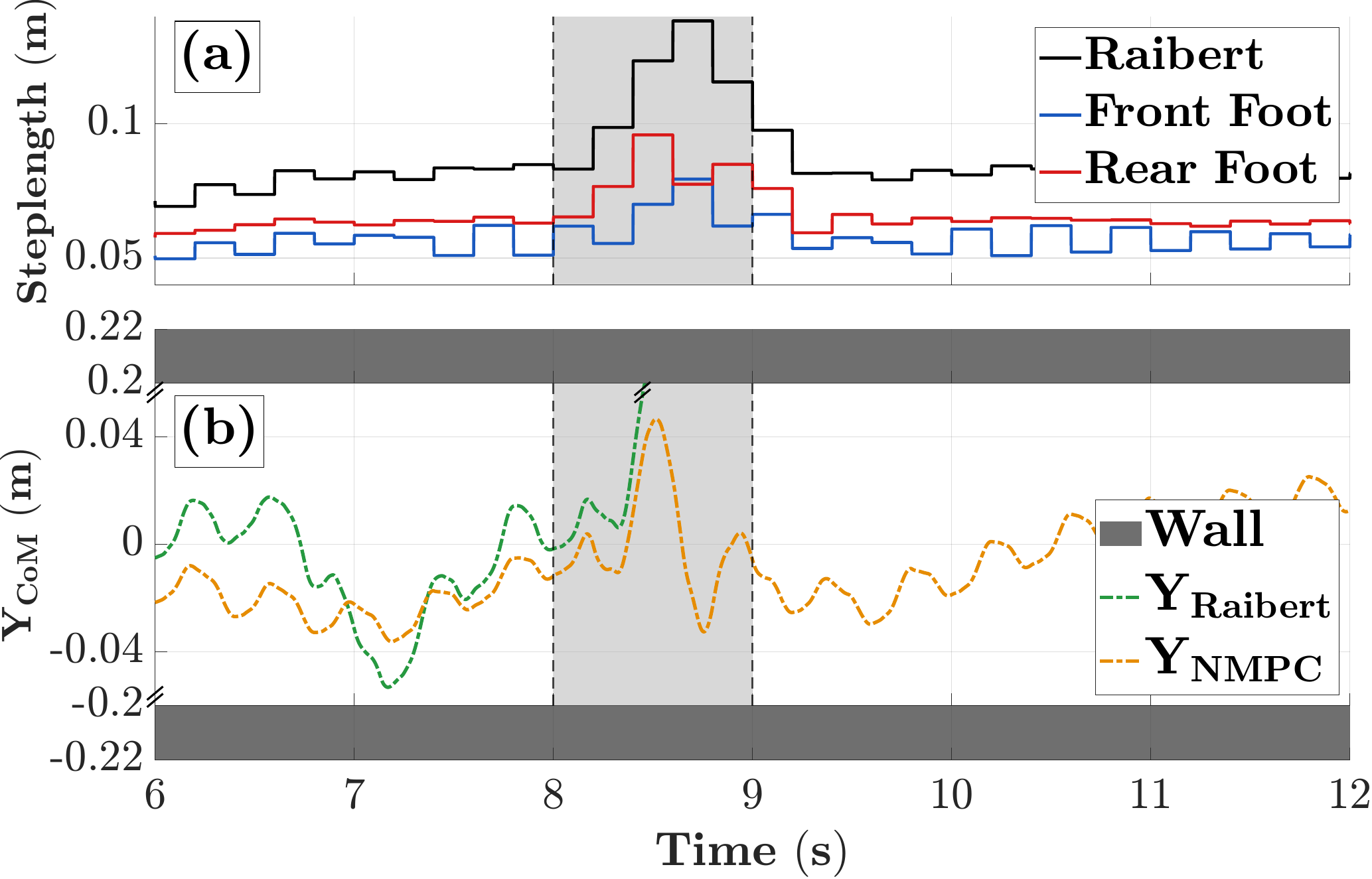}
\vspace{-0.5em}

\caption{
Plots of the (a) step length calculated by the Raibert baseline (black) and the optimum step length computed by MR-NMPC for the front foot (blue) and the rear foot (red) and (b) Lateral CoM trajectory ($y$-direction) where the Raibert heuristic (green) is deployed as a foot-step planner and the MR-NMPC is deployed as a control framework (orange). Sinusoidal force is applied as a disturbance during the shaded interval from $t=8$ s to $9$ s throughout wall-supported locomotion. The maximum magnitude of the disturbance is 50 N in the $x$-direction.}
\vspace{-1.5em}
\label{fig:push2}
\end{figure}

\subsection{Disturbance Rejection and Constraint Awareness}

\label{sec:Disturbance2}

To evaluate the robustness of the MR-NMPC against high-magnitude perturbations, we applied a $50$ N sinusoidal force from $t\!=\!8$ s to $9$ s. As shown in Fig. \ref{fig:push2} (a), the Raibert heuristic (blue) responds to the velocity spike with long step lengths that trigger joint-limit saturation, resulting in a subsequent instability as illustrated in Fig. \ref{fig:push2} (b). 
In contrast, the MR-NMPC (red) leverages its prediction horizon to identify an optimal sequence of smaller, asymmetric steps. By co-optimizing state feedback with anticipated contact timings, the proposed MR-NMPC mitigates the disturbances while respecting all safety and kinematic constraints. 

\section{Conclusions}
\label{sec:Conclusions}

This paper introduces an MR-NMPC framework that unifies high-level footstep planning and low-level ERF computation within a single optimization loop. By utilizing indicator function with SRB model, the high level multi-rate NMPC dynamically plans both the slow updating discrete-time trajectories of the contact points and the fast updating continuous-time trajectory of the SRB states. By incorporating contact-point planning within the optimal control framework, this architecture provides the flexibility to employ asymmetric step lengths and user-defined contact sequences for negotiating challenging terrain. 
The effectiveness of the proposed framework was validated through extensive simulations using the Unitree A1 platform. The results demonstrate that the MR-NMPC framework significantly enhances the robot's ability to navigate rough terrain and reject external disturbances in wall-supported configurations. Most notably, the proposed approach achieved a 2.9 times higher success rate in negotiating irregular terrain at high speeds compared to conventional MPC methods utilizing heuristic foot placement. This underscores the importance of unified trajectory and contact optimization for highly dynamic, non-standard locomotion tasks.



Preliminary hardware experiments indicate the real-time feasibility of the proposed framework, enabling the robot to transition from a crouched posture to upright wall-supported bipedal locomotion and maintain balance for over 60 s. The primary limitation is velocity estimation noise, which induces forward pitching and prevents the stable limit-cycle behavior observed in simulation. Future work will focus on improving state estimation through Kalman filtering and sensor fusion, increasing the front-foot contact area to enhance wall-support stability, and incorporating perception-aware planning to handle uncertainty in terrain geometry and contact conditions. We also plan to perform extensive hardware validation and comparative studies against stronger optimization-based baselines to further evaluate the benefits of the proposed multi-rate formulation.

\bibliographystyle{IEEEtran}
\bibliography{references}

\end{document}